\documentclass[preprint, 12pt, authoryear]{elsarticle}
\usepackage{geometry}
\usepackage{amssymb}
\usepackage{amsmath}
\usepackage{amsfonts}
\usepackage{graphicx}
\usepackage{textcomp}
\usepackage{subcaption}
\usepackage{multirow}
\usepackage{float}
\usepackage{placeins}
\usepackage{algorithm}
\usepackage{algpseudocode}
\usepackage{booktabs}
\usepackage{hyperref}
\usepackage{url}
\usepackage{tabularx}
\usepackage{ragged2e} 

\journal{Computerized Medical Imaging and Graphics}

\begin{document}

\begin{frontmatter}

\title{A-QCF-Net: An Adaptive Quaternion Cross-Fusion Network for Multimodal Liver Tumor Segmentation from Unpaired Datasets}

\author[aff1]{Arunkumar V}

\author[aff2]{V. M. Firos}

\author[aff1]{S. Senthilkumar}

\author[aff2]{G. R. Gangadharan\corref{cor1}}


\affiliation[aff1]{organization={University College of Engineering, Bharathidasan Institute of Technology Campus, Anna University},
    city={Tiruchirappalli},
    state={Tamilnadu},
    postcode={620 024},
    country={India}}

\affiliation[aff2]{organization={National Institute of Technology},
    city={Tiruchirappalli},
    state={Tamil Nadu},
    postcode={620015},
    country={India}}

\begin{abstract}
Multimodal medical imaging provides complementary information that is crucial for accurate delineation of pathology, but the development of deep learning models is limited by the scarcity of large datasets in which different modalities are paired and spatially aligned. This paper addresses this fundamental limitation by proposing an Adaptive Quaternion Cross-Fusion Network (A-QCF-Net) that learns a single unified segmentation model from completely separate and unpaired CT and MRI cohorts. The architecture exploits the parameter efficiency and expressive power of Quaternion Neural Networks to construct a shared feature space. At its core is the Adaptive Quaternion Cross-Fusion (A-QCF) block, a data driven attention module that enables bidirectional knowledge transfer between the two streams. By learning to modulate the flow of information dynamically, the A-QCF block allows the network to exchange abstract modality specific expertise, such as the sharp anatomical boundary information available in CT and the subtle soft tissue contrast provided by MRI. This mutual exchange regularizes and enriches the feature representations of both streams. We validate the framework by jointly training a single model on the unpaired LiTS (CT) and ATLAS (MRI) datasets. The jointly trained model achieves Tumor Dice scores of 76.7\% on CT and 78.3\% on MRI, significantly exceeding the strong unimodal nnU-Net baseline by margins of 5.4\% and 4.7\% respectively. Furthermore, comprehensive explainability analysis using Grad-CAM and Grad-CAM++ confirms that the model correctly focuses on relevant pathological structures, ensuring the learned representations are clinically meaningful. This provides a robust and clinically viable paradigm for unlocking the large unpaired imaging archives that are common in healthcare.
\end{abstract}


\begin{keyword}
Cross-Attention \sep Deep Learning \sep Explainable AI \sep Grad-CAM \sep Medical Image Segmentation \sep Multimodal Learning \sep Quaternion Neural Networks \sep Unpaired Data
\end{keyword}

\end{frontmatter}

\section{Introduction}
\label{sec:introduction}

Multimodal medical imaging, exemplified by Computed Tomography (CT) and Magnetic Resonance Imaging (MRI), provides complementary views that are clinically important for the diagnosis and delineation of complex pathologies such as liver tumors. In the setting of hepatic malignancy, no single imaging modality captures all clinically relevant information. Computed Tomography, with its submillimeter resolution, offers excellent anatomical detail and is often preferred for visualizing sharp organ boundaries and vascular structures \citep{heimann2009comparison}. However, its ability to distinguish subtle variations in soft tissue, for example differentiating a necrotic tumor core from a viable margin or identifying small isodense lesions, is limited. Magnetic Resonance Imaging, in contrast, excels at soft tissue contrast and can reveal tumor textures and margins that are often inconspicuous on CT \citep{gross2024automated}. Different MRI sequences emphasize distinct biophysical properties, which supports more nuanced lesion characterization. MRI is therefore superior for detecting small satellite lesions and for assessing tumor infiltration into the surrounding parenchyma. These properties suggest that a model capable of integrating the structural clarity of CT with the textural sensitivity of MRI could reach a level of precision and robustness that is not achievable with either modality alone. This clinical motivation creates a clear need for advanced multimodal AI systems.

Most existing attempts to exploit unpaired cohorts follow an indirect strategy. They first synthesize a missing modality, for example, CT to MRI or MRI to CT, and then train a segmentation network on the generated images. Such a two stage pipeline ties the final segmentation accuracy to the fidelity of the synthesis model and introduces the risk that synthesis artifacts will propagate into the predicted masks. In this work, we follow a more direct strategy: we learn segmentation from unpaired cohorts without generating synthetic images, by encouraging the network to share modality invariant semantics at the feature level during joint training. This removes the surrogate synthesis objective and aligns optimization with the end segmentation task.

Despite the clear clinical motivation for multimodal learning, the dominant deep learning paradigm for multimodal fusion is severely constrained by data availability. Most current methods assume access to large collections of paired and spatially aligned datasets, in which individual patients have undergone multiple imaging studies that are carefully coregistered \citep{stahlschmidt2022multimodal, Oktay2018}. Architectures designed for this setting, ranging from simple early fusion to more elaborate deep fusion designs, are built to exploit voxel wise correspondence between aligned scans from the same patient. However, this assumption rarely holds in routine clinical practice. Acquiring such datasets is logistically complex, expensive, and often ethically challenging, since it may require additional imaging that is not part of standard care. In contrast, most hospitals already hold large unpaired archives, for example, thousands of CT scans from one cohort and a separate collection of MRIs from another. This unpaired data setting has created a major bottleneck, since the complementary information in these archives remains largely unused.

This paper addresses the paired-data bottleneck by moving from explicit data fusion to abstract cross-modal knowledge transfer. Our central hypothesis is that a properly designed network, when jointly trained in disparate and unpaired cohorts, can learn modality-invariant semantic concepts and use this shared knowledge to mutually regularize and enhance its performance on both modalities. For example, the network can learn the general concept of “sharp anatomical boundaries” from the CT cohort and transfer this knowledge to improve organ delineation in the MRI stream. In contrast, it can learn “subtle tumor textures” from the MRI cohort to increase the sensitivity of the CT stream to isodense lesions. To achieve this, we propose an Adaptive Quaternion Cross-Fusion Network (A-QCF-Net), a novel dual-stream architecture designed for this unpaired learning task. Our primary contributions are threefold: (i) We introduce a dual-stream quaternion architecture that learns a single segmentation model from entirely unpaired CT and MRI cohorts; by using Quaternion Neural Networks (QNNs),we enforce a feature space that has parameter efficiency and is structurally entangled, that is highly conducive to learning modality-invariant representations. (ii) We propose the Adaptive QCF (A-QCF) block, a novel, dynamic cross attention mechanism that serves as the core engine for knowledge transfer; the A-QCF block learns to intelligently modulate the flow of information between the unpaired streams, a powerful cross regularizer that is driven by data, that enriches the feature space of each modality with abstract knowledge from the other. (iii) We demonstrate that this unpaired training strategy yields a single, robust model capable of segmentation that achieves high performance on either CT or MRI alone at inference time; the resulting model contains a generalized understanding of hepatic anatomy informed by both modalities, overcoming a critical barrier to the development of practical and powerful medical AI.

The remainder of this paper is structured as follows: Section \ref{sec:related_work} reviews the relevant literature. Section \ref{sec:methodology} details the proposed A-QCF-Net framework. Section \ref{sec:experiments} presents the experimental setup and results, followed by concluding remarks in Section \ref{sec:conclusion}.

\section{Related Work}
\label{sec:related_work}

The field of medical image segmentation has been shaped mainly by unimodal models, most of them derived from the U-Net architecture \citep{ronneberger2015u}. The encoder–decoder structure of U-Net has become a standard template for many later designs \citep{gross2024automated}. More recent work has extended this template with powerful Transformer based backbones such as UNETR \citep{hatamizadeh2022unetr} and a range of efficient hybrid architectures \citep{perera2024segformer3d, fu20233d} that are designed to capture long range spatial dependencies. Within the convolutional paradigm, the nnU-Net framework remains a particularly influential contribution, since it automatically configures a U-Net style pipeline for a given unimodal dataset and has set strong benchmarks across many tasks \citep{isensee2021nnu}. Although these unimodal systems are highly effective, they are inherently limited by the information content of a single modality. Such models, which we refer to as data silo models, have no mechanism to access or exploit complementary information that is present in other imaging modalities. A network trained only on CT cannot learn the subtle soft tissue patterns that are visible only in MRI. No amount of additional unimodal data or computational capacity can compensate for the lack of these modality specific cues. As a result, there is an implicit upper bound on performance that is imposed by the single modality itself. Moving beyond this ceiling requires a shift from unimodal design toward models that can integrate information across multiple modalities in a principled way.

Most existing approaches to multimodal integration are built around the requirement that datasets provide paired and spatially aligned images. Within this setting, a wide range of fusion strategies has been investigated, from simple concatenation at the input level to more sophisticated deep fusion schemes. These include explicit information interaction modules \citep{fan2024multimodal}, Transformer based fusion architectures \citep{ni2024tran}, and models that support multiple tasks within a unified framework \citep{marinov2023mirror, gui2025s2net}. These methods clearly show that multimodal fusion can improve performance when well curated paired data are available.

However, the dependence on carefully aligned paired datasets creates a significant gap between research practice and clinical reality. In many published studies, the data come from rare highly curated cohorts that are difficult to assemble and maintain. In typical clinical settings, however, such collections are the exception rather than the norm. Acquiring large paired datasets is logistically demanding, financially expensive, and often ethically challenging, since it may require additional imaging that is not part of standard care. In routine practice, institutions are more likely to possess large independent archives, for example separate cohorts of CT and MRI scans that share no patient level alignment. As a consequence, a large body of multimodal work has limited direct translational potential. 

There is therefore a fundamental need for methods that can realise the benefits of cross modal learning without relying on strict pairing or voxel wise alignment. Our work addresses this gap by reframing cross modal interaction. Instead of treating fusion as patient specific feature combination, we view it as a mechanism for abstract knowledge transfer between cohorts that are entirely unpaired.

The problem of learning from unpaired data has given rise to two main research directions, namely unpaired image synthesis and direct cross modal learning. The first and more widely explored direction focuses on generating a synthetic version of the missing modality. Early work in this area was driven by Generative Adversarial Networks such as CycleGAN \citep{zhu2017unpaired}. More recently, high fidelity diffusion models have become the method of choice for unpaired synthesis, with approaches that emphasize structural consistency \citep{chen2024contourdiff}, cross modal conditioning \citep{xing2024cross}, or improved alignment for specific applications such as liver tumor segmentation \citep{chen2024diff4mmlits}. This line of work has led to label free frameworks in which segmentation networks are trained entirely on synthetic images \citep{hu2023label}. However, this remains an indirect two stage strategy, as reviewed in \citep{qiegen2025diffusion}, in which the final segmentation performance is tightly constrained by the quality of the synthetic images. Any artifact or hallucinated structure in the generated modality can be propagated and amplified by the segmentation network, and several studies have shown that even very high quality synthesis may still limit downstream performance compared to training directly on real data \citep{graf2023denoising}. The second paradigm is Unsupervised Domain Adaptation, where a model is adapted from a labeled source modality to an unlabeled target modality using adversarial or contrastive objectives \citep{hong2022unsupervised, chen2025unsupervised}. Other direct strategies include collaborative learning between multiple streams \citep{liu2023modality}, representations that disentangle modality specific style from shared anatomical content \citep{chartsias2020disentangle}, and knowledge distillation between networks \citep{dou2020unpaired}. Although these methods are effective, they typically assume a directional transfer from source to target and do not aim to learn a single model that is trained symmetrically on both modalities. Our framework builds on this direct family of approaches and introduces a bidirectional cross fusion mechanism within a unified quaternion architecture, with the explicit goal of learning one shared model for both CT and MRI.

\section{Methodology}
\label{sec:methodology}

\begin{figure}[htbp]
    \centering
    \includegraphics[width=0.9\textwidth]{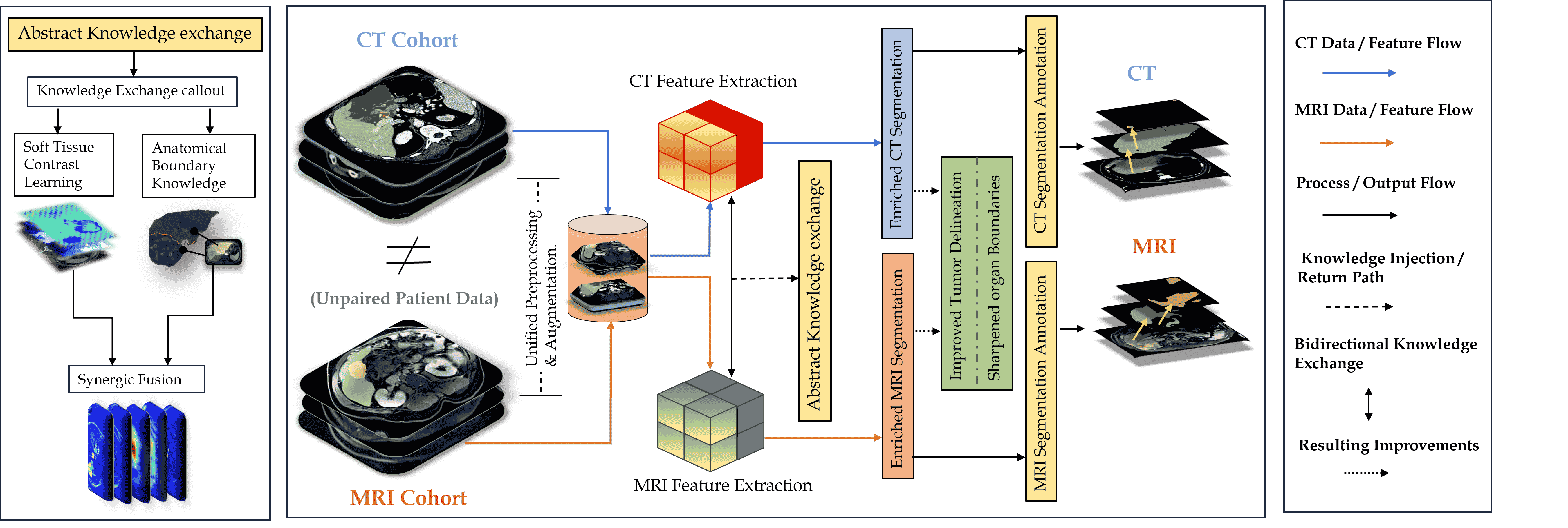}
    \caption{The conceptual framework of our unpaired cross-modal learning approach. Data from separate, unpaired CT and MRI cohorts are processed in parallel. An abstract knowledge exchange mechanism allows the network to transfer generalized features (e.g., sharp boundary information from CT, soft-tissue texture from MRI) between the streams. This enriches the segmentation capability of both modalities, leading to improved tumor delineation and more accurate annotations from a single, jointly-trained model.}
    \label{fig:conceptual_framework}
\end{figure}

The proposed A-QCF-Net methodology is a symmetric dual stream encoder–decoder network that is tailored to our unpaired training setting. As illustrated in Figure~\ref{fig:conceptual_framework}, the core idea is to process, in a single training step, a batch that contains two completely unrelated images, one CT scan and one MRI scan, in parallel. At each level of the twin encoders, feature representations are exchanged and dynamically regularized by the proposed Adaptive Quaternion Cross-Fusion (A-QCF) block. After this multiscale knowledge transfer, both streams pass through a shared quaternion bottleneck, which is a key architectural choice that encourages the learning of a common, highly abstract feature space.

The two decoder pathways then reconstruct the segmentation map for each modality independently. They use skip connections that are filtered by attention gates to restore fine spatial detail. A joint loss, computed as the sum of the losses from the CT and MRI streams, drives the optimization. This training strategy forces the network to learn a single unified set of weights that can segment both modalities effectively. The final model is therefore a single network that can be deployed flexibly for unimodal inference on either CT or MRI data.

\subsection{Unpaired Multimodal Data Pipeline}
A dedicated data pipeline is required to support the unpaired training scheme. Its role is to load data from different sources, apply modality specific preprocessing, perform robust data augmentation, and assemble batches of unrelated samples for joint optimization.

\subsubsection{Datasets and Task Definition}
We build our training and validation cohorts from two publicly available independent datasets. The CT cohort is drawn from the Liver Tumor Segmentation Challenge (LiTS) dataset \citep{bilic2023liver}, and the MRI cohort consists of T1 weighted scans from the ATLAS dataset \citep{quinton2023tumour}. There is no patient level overlap between the two datasets. For both LiTS and ATLAS, we employ a five fold cross validation protocol. The cohorts are divided into five distinct folds at the patient level, stratified by tumor presence to maintain a comparable proportion of positive cases across folds. The task is a three class semantic segmentation at the voxel level with the following labels: Class 0 (background), Class 1 (liver), and Class 2 (tumor).

\subsubsection{Modality Specific Preprocessing}
We apply a custom preprocessing pipeline with distinct orientation and intensity normalization steps for each modality, to harmonize inputs from the heterogeneous sources. For CT scans, we convert the data to an RAS orientation and normalize voxel intensities from a Hounsfield Unit (HU) window of $[-16, 176]$ to a floating point range of $[0.0, 1.0]$. For MRI scans, we use an LAS orientation and scale intensities from $[74, 511]$ to $[0.0, 1.0]$. For an input image $I$, the complete preprocessing pipeline is a composition of standard transforms: isotropic resampling to a $(1, 1, 1)$ mm spacing, padding so that spatial dimensions are divisible by 16, and foreground cropping to remove excess background and concentrate computation on the relevant anatomy.

\subsubsection{Data Augmentation and Unpaired Batch Construction}
To improve generalization, we apply an extensive data augmentation scheme independently to each modality. The augmentations include random flips, $90^{\circ}$ rotations, and random intensity shifts. During training, we extract patches of size $(256, 256, 16)$ with a sampling strategy that favours patches containing foreground voxels, so that the network sees relevant anatomical structures frequently.

For each training iteration, the dataloader independently samples one CT patch and one MRI patch and combines them into a single batch. As a result, the CT and MRI samples processed together in a given step come from different patients and different datasets. This construction enforces the unpaired nature of the training and ensures that any cross modal interaction learned by the model arises from shared semantics rather than from direct voxel wise correspondence.

The cornerstone of our unpaired learning strategy is the batch construction. At each training iteration $t$, the dataloader produces a batch $B_t$ containing one random fully-augmented sample from the CT dataset and one from the MRI dataset:
\begin{equation}
    B_t = \{ (x_{ct}^{(i)}, y_{ct}^{(i)}), (x_{mri}^{(j)}, y_{mri}^{(j)}) \}
\end{equation}
where subject index $i$ from the CT cohort is independent of subject index $j$ from the MRI cohort. A custom collate function assembles these two unrelated samples into a single batch, ensuring the network is trained exclusively on unpaired cross-modal data.

\subsection{Adaptive Quaternion Cross-Fusion Network (A-QCF-Net) Architecture}
\label{sec:qcf_architecture}

\begin{figure}[htbp]
    \centering
    \includegraphics[width=0.95\textwidth]{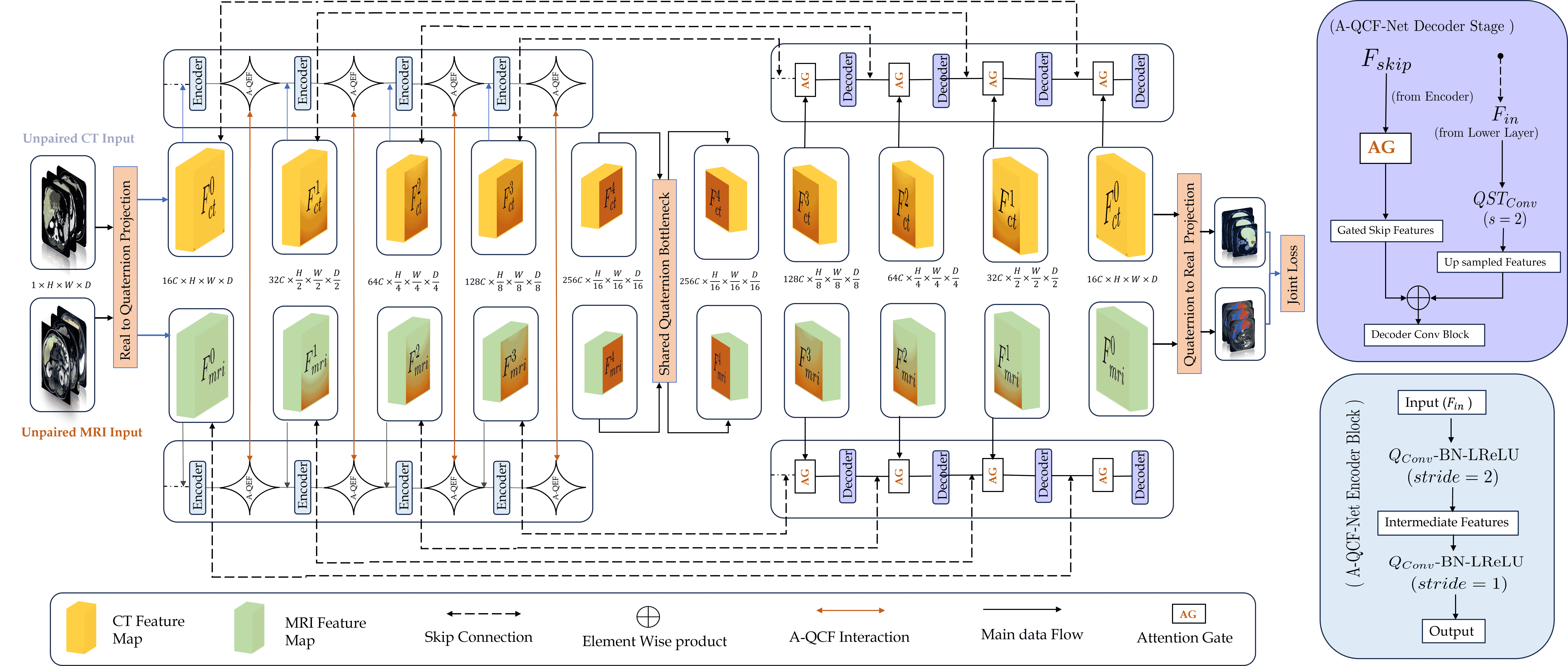} 
    \caption{The detailed architecture of the A-QCF-Net. Unpaired CT and MRI inputs are processed by parallel quaternion encoders. At each scale, the Adaptive QCF (A-QCF) blocks (detailed in Fig.~\ref{fig:aqcf_diagram}) facilitate dynamic knowledge transfer. Both streams converge into a shared quaternion bottleneck to learn a common feature representation. The decoders use attention-gated (AG) skip connections to reconstruct the segmentation outputs. A joint loss drives the optimization.}
    \label{fig:architecture_diagram}
\end{figure}

The core of our framework is the A-QCF-Net, a novel dual-stream network built on quaternion algebra to enable adaptive knowledge transfer from unpaired data. The high-level architecture, depicted in Figure~\ref{fig:architecture_diagram}, integrates three primary components: quaternion convolutional layers as efficient feature extractors, the novel A-QCF block for dynamic inter-stream communication, and a symmetric dual-stream encoder-decoder structure with a shared bottleneck.

\subsubsection*{Notation and tensor shapes}
Unless noted otherwise, all feature tensors are 3D with a channel first layout. We use $B$ for batch size, $C$ for (quaternion) channel width per stream, and $(D,H,W)$ for depth, height and width.

\begin{table}[H]
\centering
\caption{Summary of notation and tensor shapes used in the A-QCF-Net architecture.}
\label{tab:notation}
\footnotesize
\begin{tabular}{lll}
\toprule
\textbf{Symbol} & \textbf{Description} & \textbf{Shape} \\
\midrule
$F_{ct}$, $F_{mri}$ & encoder features at a given scale (CT/MRI) & $\mathbb{R}^{B\times 4C\times D\times H\times W}$ \\
$Q$, $K$, $V$ & query, key, value after $1\!\times\!1\!\times\!1$ QConv & $\mathbb{R}^{B\times 4C\times D\times H\times W}$ \\
$\mathcal{A}$ & channel-attention map (per voxel) & $\mathbb{R}^{B\times 4C\times D\times H\times W}$ \\
$C_{\text{raw}}$ & raw cross-context (value modulated by attention) & $\mathbb{R}^{B\times 4C\times D\times H\times W}$ \\
$\lambda$ & adaptive gate (broadcast over $D, H, W$) & $\mathbb{R}^{B\times 1\times 1\times 1\times 1}$ \\
\bottomrule
\end{tabular}
\end{table}

\noindent\textbf{Concat axis:} We concatenate along channels ($\mathrm{axis}=1$). \textbf{Softmax axis:} Channel-wise ($\mathrm{axis}=1$) at each spatial location $(d,h,w)$.

\subsubsection{Quaternion Convolution as the Foundational Layer}

The A-QCF-Net utilizes Quaternion Neural Networks (QNNs) to construct a shared feature space that is both parameter efficient and structurally entangled. Originally proposed for speech processing \citep{Parcollet2018} and later adapted for vision \citep{zhu2018quaternion}, QNNs process data in the hypercomplex domain. Quaternion representations have been successfully used for multimodal fusion in paired settings \citep{deng2023qmls}, but their use as a backbone for direct knowledge transfer between unpaired modalities remains unexplored. Recent studies highlight the potential of quaternion networks for domain generalization \citep{sigillo2025generalizing} and self attention mechanisms \citep{zhan2025cosegnet}. These findings support our hypothesis that QNNs are a natural choice for constructing the compact, shared feature space required to bridge large, siloed CT and MRI archives.

A quaternion is a hypercomplex number defined as $q = a + b\mathbf{i} + c\mathbf{j} + d\mathbf{k}$. In our architecture, a real valued feature map with $4C$ channels is re-interpreted as a quaternion valued map with $C$ channels. The convolution operation is governed by the non commutative Hamilton product. For an input tensor $X$ and a quaternion kernel $W = W_r + W_i\mathbf{i} + W_j\mathbf{j} + W_k\mathbf{k}$, this operation is realized by constructing a structured real valued kernel matrix, $W_{\text{real}}$, that simulates the hypercomplex multiplication \citep{zhu2018quaternion}:
\begin{equation}
    W_{\text{real}} = 
    \begin{pmatrix}
        W_r & -W_i & -W_j & -W_k \\
        W_i & W_r & -W_k & W_j \\
        W_j & W_k & W_r & -W_i \\
        W_k & -W_j & W_i & W_r 
    \end{pmatrix}
\end{equation}
This formulation offers two specific advantages for our unpaired learning task: (i) The algebraic structure of the Hamilton product imposes fixed ties between sub-kernels. Unlike a standard real valued convolution where every filter is independent, the quaternion structure acts as an inductive bias that encourages the learning of correlated and holistic features (such as the relationship between edges and textures) rather than independent filters. (ii) It is highly parameter efficient \citep{wang2023qgd}. Consider a kernel size as $k$. A standard real valued 3D convolution mapping $4C_{\text{in}}$ to $4C_{\text{out}}$ channels requires a parameter count of $\theta_{\text{real}} = 16\,C_{\text{out}}C_{\text{in}}k^3$. In contrast, a quaternion convolution relies only on the four sub-kernels $W_r,W_i,W_j,W_k$, resulting in:
\begin{equation}
\#\theta_{\mathbb{H}} = 4\,C_{\text{out}}C_{\text{in}}k^3, \quad \text{yielding} \quad \frac{\#\theta_{\mathbb{H}}}{\#\theta_{\text{real}}} = \frac{1}{4}
\end{equation}
This fourfold reduction in parameters is especially helpful when training a single model across disparate data distributions, such as our unpaired CT and MRI cohorts, since it reduces the risk of overfitting while maintaining feature expressiveness.

\subsubsection{The Adaptive Quaternion Cross-Fusion (A-QCF) Block}

Classical channel-attention mechanisms reweight channels within a \emph{single} stream by summarizing features (e.g., global average pooling) and predicting per-channel gains via a light MLP, while spatial attention emphasizes salient regions. Attention U-Net~\citep{Oktay2018} brings this idea to encoder–decoder segmentation by gating \emph{skip} features using a deeper context, again within one modality. Our Adaptive QCF gate differs in two ways: (i) It is \emph{cross-modal}: the query originates from the target stream (e.g., CT) while key/value come from the source stream (e.g., MRI), enabling knowledge transfer rather than mere self-reweighting. (ii) We perform a \emph{channel-wise softmax} over $Q\odot K$, which normalizes competitive evidence across channels at each spatial location, instead of collapsing spatial structure with a single global descriptor. The gate $\lambda\in[0,1]$ is then conditioned on pooled statistics from both the original features and the offered cross-context, allowing the network to admit or suppress foreign information in a data-driven manner.

\begin{figure} 
    \centering
    \includegraphics[width=\columnwidth]{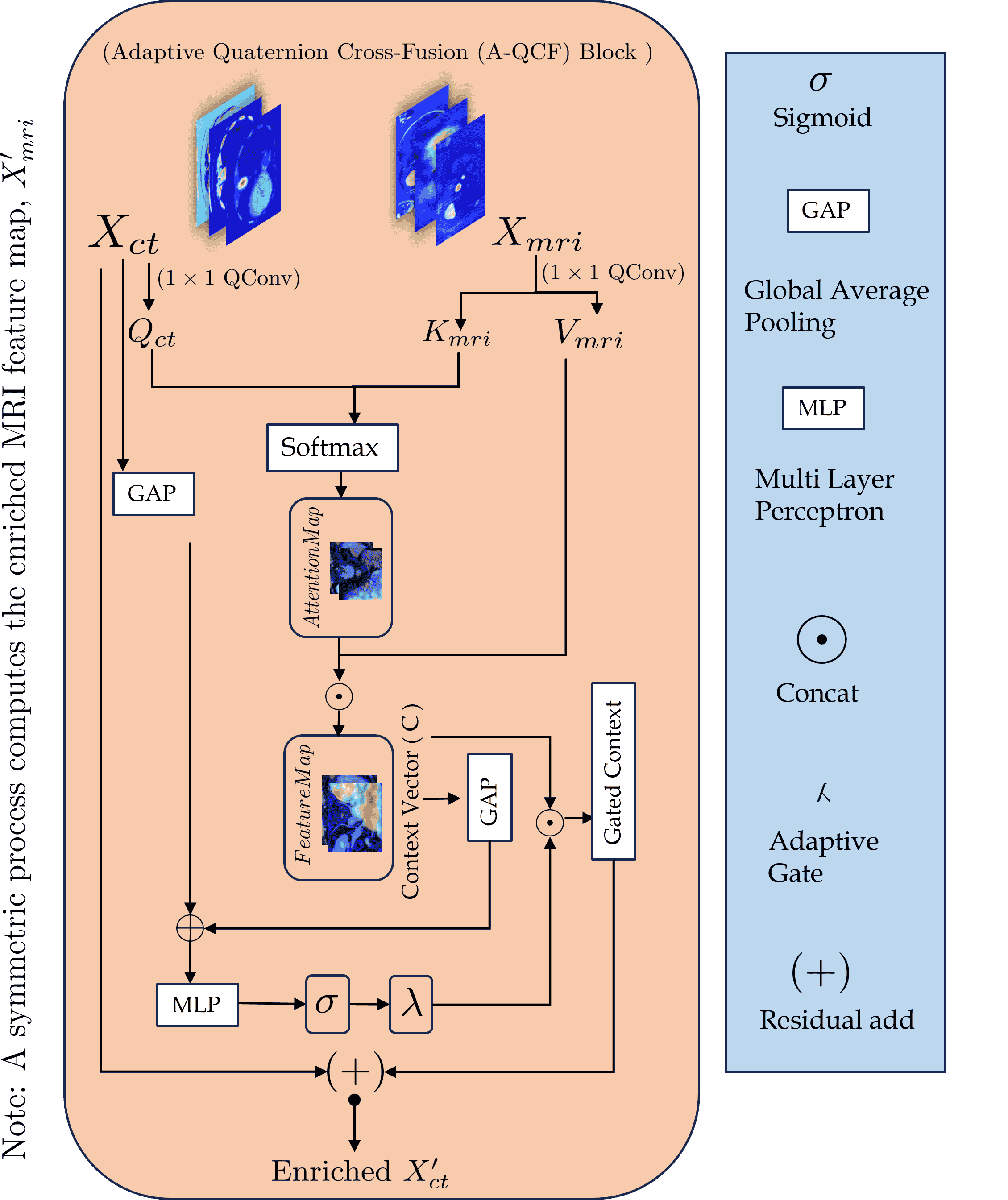} 
    \caption{Detailed visualization of the Adaptive Quaternion Cross-Fusion (A-QCF) block for the MRI-to-CT knowledge transfer path. The block generates a context vector from the MRI stream, modulates it with a data-driven adaptive gate ($\lambda$), and fuses the resulting gated context with the CT feature map to produce an enriched representation ($X'_{ct}$). The process occurs symmetrically for the CT-to-MRI path.}
    \label{fig:aqcf_diagram}
\end{figure}

The primary mechanism for cross-modal learning is our proposed A-QCF block, the data flow of which is visualized in Figure~\ref{fig:aqcf_diagram}. Positioned between the parallel encoder streams, it facilitates dynamic bidirectional transfer of abstract knowledge. Since the input feature maps $F_{ct}$ and $F_{mri}$ are from different patients, the block learns to exchange generalized feature patterns rather than patient-specific spatial correlations. This is achieved through a bespoke attention mechanism augmented with a gating module that is driven by the data. The process is detailed in Algorithm~\ref{alg:aqcf} and involves the following five steps for the MRI-to-CT knowledge transfer path.

\vspace{0.5em}
\noindent\textbf{1) Quaternion Projection:} As shown in  Figure~\ref{fig:aqcf_diagram}, input features are projected into the Query ($Q$), Key ($K$), and Value ($V$) spaces using 1x1x1 quaternion convolutions ($\text{QConv}_{1\times1\times1}$).
\begin{gather}
    Q_{ct} = \text{QConv}_{q}(F_{ct}); \quad K_{mri} = \text{QConv}_{k}(F_{mri}); \nonumber \\
    V_{mri} = \text{QConv}_{v}(F_{mri})
\end{gather}

\vspace{0.5em}
\noindent\textbf{2) Channel-wise Attention:} Attention weights are computed via an element-wise product ($\odot$) between the query and key, followed by a softmax activation along the channel dimension.
\begin{equation}
    \mathcal{A}_{mri \to ct} = \text{softmax}(Q_{ct} \odot K_{mri}, \text{dim=channel})
\end{equation}

\vspace{0.5em}
\noindent\textbf{3) Context Vector Generation:} The attention map is applied to the value tensor to generate a raw context vector $C_{raw}$ representing the abstracted knowledge from the MRI domain.
\begin{equation}
    C_{raw} = \mathcal{A}_{mri \to ct} \odot V_{mri}
\end{equation}

\vspace{0.5em}
\noindent\textbf{4) Adaptive Gating Mechanism:} To make the knowledge transfer dynamic, we introduce a learnable gate $\lambda \in [0, 1]$. This gate is generated by a small sub-network that assesses the relationship between the original features ($F_{ct}$) and the offered context ($C_{raw}$). Both tensors are summarized using Global Average Pooling (GAP), concatenated, and processed by a small MLP with a final Sigmoid activation ($\sigma$) to produce the gate value.
\begin{gather}
    s_{ct} = \text{GAP}(F_{ct}); \quad s_{context} = \text{GAP}(C_{raw}) \\
    \lambda_{mri \to ct} = \sigma(\text{MLP}(\text{concat}[s_{ct}, s_{context}]))
\end{gather}

\vspace{0.5em}
\noindent\textbf{5) Gated Feature Fusion:} The final gated context vector is created by multiplying the gate with the raw context. This is then concatenated with the original feature map and fused via a 3x3x3 quaternion convolution block ($\text{QConv}_{\text{out}}$) to produce the updated regularized feature map, $F'_{ct}$.
\begin{equation}
    F'_{ct} = \text{QConv}_{\text{out}}(\text{concat}[F_{ct}, \lambda_{mri \to ct} \cdot C_{raw}])
\end{equation}
This entire five-step process occurs symmetrically and simultaneously to compute an updated MRI feature map, $F'_{mri}$, thus completing the adaptive bidirectional knowledge transfer.

\FloatBarrier

\begin{algorithm}[htbp]
\caption{The Adaptive QCF (A-QCF) Block}\label{alg:aqcf}
\begin{algorithmic}[1]
    \State \textbf{Input:} Unpaired feature maps $F_{ct}, F_{mri}$.
    \State \textbf{Output:} Fused feature maps $F'_{ct}, F'_{mri}$.
    \Statex
    
    \Function{ComputeContext}{$F_{query}, F_{key\_val}$}
        \State $Q \leftarrow \text{QConv}_{q}(F_{query})$
        \State $K \leftarrow \text{QConv}_{k}(F_{key\_val})$; $V \leftarrow \text{QConv}_{v}(F_{key\_val})$
        \State $\mathcal{A} \leftarrow \text{softmax}(Q \odot K, \text{dim=channel})$
        \State $C_{raw} \leftarrow \mathcal{A} \odot V$
        \State \textbf{return} $C_{raw}$
    \EndFunction
    \Statex
    
    \Function{ComputeGate}{$F_{original}, C_{raw}$}
        \State $s_{orig} \leftarrow \text{GAP}(F_{original})$; $s_{ctx} \leftarrow \text{GAP}(C_{raw})$
        \State $s_{comb} \leftarrow \text{concat}[s_{orig}, s_{ctx}]$
        \State $\lambda \leftarrow \sigma(\text{MLP}(s_{comb}))$
        \State \textbf{return} $\lambda$
    \EndFunction
    \Statex
    
    \State // MRI to CT Transfer
    \State $C_{mri \to ct} \leftarrow \text{ComputeContext}(F_{ct}, F_{mri})$
    \State $\lambda_{mri \to ct} \leftarrow \text{ComputeGate}(F_{ct}, C_{mri \to ct})$
    \State $F'_{ct} \leftarrow \text{QConv}_{\text{out}}(\text{concat}[F_{ct}, \lambda_{mri \to ct} \cdot C_{mri \to ct}])$
    \Statex
    
    \State // CT to MRI Transfer (Symmetric)
    \State $C_{ct \to mri} \leftarrow \text{ComputeContext}(F_{mri}, F_{ct})$
    \State $\lambda_{ct \to mri} \leftarrow \text{ComputeGate}(F_{mri}, C_{ct \to mri})$
    \State $F'_{mri} \leftarrow \text{QConv}_{\text{out}}(\text{concat}[F_{mri}, \lambda_{ct \to mri} \cdot C_{ct \to mri}])$
    \Statex
    
    \State \textbf{return} $F'_{ct}, F'_{mri}$
\end{algorithmic}
\end{algorithm}

\FloatBarrier

A critical property for deployment is the behavior of the model when only one modality is present. If one input stream is zeroed (e.g., $F_{mri}\equiv 0$), then its projected Key and Value tensors are also zero ($K_{mri}=V_{mri}\equiv 0$), causing the context vector to collapse ($C_{raw}\equiv 0$). The update rule for the active stream gracefully degrades to a residual connection:
\begin{equation}
    F'_{ct} = \text{QConv}_{\text{out}}(\text{concat}[F_{ct}, 0]).
    \label{eq:unimodal_degrade}
\end{equation}
This ensures a stable and predictable unimodal forward pass without architectural changes.

\textit{Lemma (Lipschitz stability).} If the fusion block $\text{QConv}_{\text{out}}$ is $L$-Lipschitz in its second (concatenated) argument and $0 \le \lambda \le 1$, then the deviation from the unimodal path is bounded by the gated cross-context magnitude:
\begin{equation}
    \big\| F'_{ct} - \text{QConv}_{\text{out}}([F_{ct}\|\;0]) \big\| \le L\,\big\|\lambda\odot C_{raw}\big\|.
    \label{eq:lipschitz_bound}
\end{equation}

\subsubsection{Dual-Stream Encoder-Decoder Structure}
The overall A-QCF-Net follows a U-Net-like architecture with two parallel encoder pathways, as shown in Fig.~\ref{fig:architecture_diagram}. Each encoder consists of four down-sampling blocks, with quaternion channel counts progressing through (12, 24, 48, 96, 192,256). After each down-sampling stage, an A-QCF block is applied to facilitate knowledge transfer across multiple scales. Following the final encoder stage, both feature streams are passed through a shared quaternion bottleneck. This forces both streams through a common set of weights at the deepest, most abstract level, compelling the model to learn a compact, modality-agnostic representation.

The decoder mirrors this structure with two parallel up-sampling pathways. To recover highly granular spatial details, skip connections from the corresponding encoder stage are utilized. Each skip connection is passed through a quaternion attention gate \citep{Oktay2018}, which uses the feature map from the deeper layer to re-weight the skip connection features, allowing the model to focus on the most salient information. The final layers of each decoder stream use a 1x1x1 real-valued convolution to project the quaternion feature maps back to the real domain, producing the segmentation output with three classes for each modality.

\subsection{Joint Training and Unimodal Inference Strategy}
\label{sec:training_inference}

The A-QCF-Net is trained using a joint optimization strategy on the unpaired data streams and is uniquely designed for a flexible unimodal inference process.

\subsubsection{Joint Loss Function}
The model is trained from end to end by minimizing a joint loss function. For each forward pass, the model produces two separate segmentation outputs: a prediction for the CT input $\hat{y}_{ct}$ and a prediction for the unrelated MRI input $\hat{y}_{mri}$. The loss for each output is computed independently against its respective ground truth label ($y_{ct}$, $y_{mri}$). The total loss for a single training step, $L_{\text{total}}$, is the unweighted sum of these two independent losses:
\begin{equation}
    L_{\text{total}} = \mathcal{L}_{DiceCE}(y_{ct}, \hat{y}_{ct}) + \mathcal{L}_{DiceCE}(y_{mri}, \hat{y}_{mri})
    \label{eq:loss}
\end{equation}
This joint objective compels the network to learn a unified set of weights that can successfully segment both modalities. The A-QCF block acts as the critical information conduit, allowing the gradients from one stream to influence and regularize the feature representations of the other, thereby forcing the model to converge on a shared, modality-invariant feature space.

\subsubsection*{Segmentation Loss Components}
Let $p_{n,c}\in[0,1]$ be the predicted probability for voxel $n$ and class $c$, and $y_{n,c}\in\{0,1\}$ the one-hot label over $C$ classes. With $N$ voxels, the per-class Soft Dice is:
\begin{equation}
\mathrm{Dice}_c \;=\; \frac{2\sum_{n=1}^{N} p_{n,c}\,y_{n,c} + \varepsilon}{\sum_{n=1}^{N} p_{n,c} \;+\; \sum_{n=1}^{N} y_{n,c} \;+\; \varepsilon},
\end{equation}
and the Dice loss is the complement of the mean Dice score across classes:
\begin{equation}
\mathcal{L}_{\mathrm{Dice}} \;=\; 1 \;-\; \frac{1}{C}\sum_{c=1}^{C} \mathrm{Dice}_c.
\end{equation}
The multi-class cross-entropy is:
\begin{equation}
\mathcal{L}_{\mathrm{CE}} \;=\; -\,\frac{1}{N}\sum_{n=1}^{N}\sum_{c=1}^{C} y_{n,c}\,\log(p_{n,c}).
\end{equation}
We use the standard unweighted combination $\mathcal{L}_{\mathrm{DiceCE}} = \mathcal{L}_{\mathrm{Dice}} + \mathcal{L}_{\mathrm{CE}}$.

\subsubsection{Optimization and Training Procedure}

We optimize the model using AdamW (Adam with decoupled weight decay) \citep{zhang2023provable} for 200 epochs. The update for parameters $\theta$ at step $t$ with learning rate $\eta$ and gradients $g_t$ is given by:
\begin{equation}
\begin{aligned}
    m_t &= \beta_1 m_{t-1}+(1{-}\beta_1)g_t, \\
    v_t &= \beta_2 v_{t-1}+(1{-}\beta_2)g_t^2, \\
    \theta_{t+1} &= \theta_t-\eta\left(\frac{m_t/(1-\beta_1^t)}{\sqrt{v_t/(1-\beta_2^t)}+\epsilon}+\lambda_{\mathrm{wd}}\theta_t\right),
\end{aligned}
\label{eq:adam_update}
\end{equation}
with an initial learning rate $\eta=1 \times 10^{-4}$ and weight decay $\lambda_{\mathrm{wd}}=1 \times 10^{-5}$. A ReduceLR-OnPlateau learning rate scheduler monitors the average validation Dice score across both CT and MRI sets. If the average score does not improve for 4 consecutive epochs, the learning rate is halved. This complete training process is presented in Algorithm~\ref{alg:unpaired_step}.

\begin{algorithm}[H]
\caption{Unpaired Joint Training Step}\label{alg:unpaired_step}
\begin{algorithmic}[1]
\Require CT dataset $\mathcal{D}_{ct}$, MRI dataset $\mathcal{D}_{mri}$, model $f_{\theta}$, optimizer $\mathcal{O}$
\State Sample $(x_{ct},y_{ct})\sim\mathcal{D}_{ct}$ and $(x_{mri},y_{mri})\sim\mathcal{D}_{mri}$ independently.
\State Apply modality-specific preprocessing and augmentation to each sample.
\State $(\hat{y}_{ct},\hat{y}_{mri}) \leftarrow f_{\theta}(x_{ct},x_{mri})$ \Comment{Forward pass with A-QCF at each scale}
\State $\ell_{ct}\leftarrow \mathcal{L}_{\mathrm{DiceCE}}(y_{ct},\hat{y}_{ct})$; \quad $\ell_{mri}\leftarrow \mathcal{L}_{\mathrm{DiceCE}}(y_{mri},\hat{y}_{mri})$
\State $\ell_{\text{total}}\leftarrow \ell_{ct}+\ell_{mri}$ \Comment{Sum losses from independent streams}
\State $\mathcal{O}.\mathrm{zero\_grad}()$;\; $\ell_{\text{total}}.\mathrm{backward}()$;\; $\mathcal{O}.\mathrm{step}()$ \Comment{Update shared weights $\theta$}
\end{algorithmic}
\end{algorithm}

\subsubsection{Flexible Unimodal Inference}
A key property of the framework is its robust and flexible behaviour at inference time. For full 3D volumes, we use sliding window inference with a patch size of $256 \times 256 \times 16$, an overlap of 0.8, and Gaussian blending to obtain smooth predictions. At test time, the model receives a single input volume $x_{in}$ together with a modality flag (CT or MRI). The stream corresponding to the absent modality is given a zero tensor of identical shape. For example, if the input is a CT scan, we set
\begin{equation}
    x_{\text{stream, CT}} = x_{in}, \quad x_{\text{stream, MRI}} = \mathbf{0}.
\end{equation}
Both the real input and the zero tensor are then passed through their respective encoder streams, so that the computation graph is identical to the one used during training. As discussed earlier, when one stream receives only zeros, the associated cross context from the A-QCF blocks collapses. The result is a stable unimodal forward pass that still makes use of all learned weights.

\section{Experiments and Results}
\label{sec:experiments}

To evaluate the proposed A-QCF-Net framework, we performed a comprehensive set of experiments with four main aims: (1) to quantitatively compare our unpaired cross modal training strategy with strong unimodal and state of the art baselines, (2) to provide qualitative evidence of segmentation accuracy, (3) to assess the trustworthiness of the model through explainability analysis, and (4) to analyse the contribution of the main architectural components by means of ablation studies.

\subsection{Experimental Setup}

\subsubsection{Evaluation Protocol}
To obtain a robust and unbiased assessment, all models were trained and validated using a five fold cross validation protocol. The patient cohorts in both the LiTS (CT) and ATLAS (MRI) datasets were split into five distinct, non overlapping folds at the patient level. In each iteration, one fold was held out for testing and the remaining four folds were used for training. All reported performance values are given as mean $\pm$ standard deviation across the five folds to reflect both central tendency and variability. Segmentation performance was quantified using the Dice Similarity Coefficient (DSC) for volumetric overlap, the 95\% Hausdorff Distance (HD95, in mm) for boundary agreement, and the Mean Surface Distance (MSD, in mm) to assess average surface deviation. To formally test performance differences, we conducted a statistical analysis based on the non parametric Wilcoxon signed rank test applied to the fold wise scores. A Bonferroni correction was applied to the significance threshold ($\alpha = 0.05$) to account for multiple comparisons.

\subsubsection{Clinical Validation Study Design}
To evaluate the clinical acceptability and practical value of A-QCF-Net, we carried out a retrospective reader study with two board certified abdominal radiologists (R1 and R2), each with more than 10 years of experience. We selected a stratified random sample of 40 held out test cases, balanced by imaging modality (20 CT and 20 MRI), lesion size, and boundary clarity.

The readers independently reviewed each case in randomized order to reduce potential bias. For each case, they were shown the original scan together with the corresponding segmentation mask generated by A-QCF-Net. They were asked to rate the segmentation quality on a 10 point Likert scale, where 1 corresponds to poor or unusable quality and 10 corresponds to excellent or near perfect quality. The primary endpoint of the study was the mean quality score. Secondary endpoints included the proportion of cases judged clinically acceptable, defined as segmentation that would not require no edits or only minor edits for use in treatment planning.

To measure the consistency between the two readers, we computed the Intraclass Correlation Coefficient (ICC, two way random, absolute agreement) for the continuous Likert scores and Cohen's Kappa ($\kappa$) for the binary clinical acceptability labels.

\subsubsection{Baselines for Comparison} We interpret the performance of A-QCF-Net in relation to a broad set of strong baseline methods, grouped into three categories. 1) Primary Unimodal Benchmarks: Unimodal nnU-Net \citep{isensee2021nnu} and Attention U-Net \citep{Oktay2018}. 2) Advanced Segmentation Architectures: Transformer-based unimodal models including UNETR \citep{hatamizadeh2022unetr}, Swin UNETR \citep{hatamizadeh2022swin}, and other hybrids \citep{ou2024restransunet, chen2022efficient}. 3) Competing Unpaired Learning Methods: Frameworks based on knowledge distillation (Unpaired via KD \citep{dou2020unpaired}), adversarial learning (Cross-Modal Adv. \citep{ozkan2024cross}), collaborative learning (MCTHNet \citep{liu2023modality}), and image synthesis (Diff4MMLITS \citep{chen2024diff4mmlits}). For a fair comparison, all baselines were trained using identical data folds, preprocessing, and augmentation.

\subsubsection{Implementation Details}
The A-QCF-Net framework was implemented in PyTorch and MONAI. The model was trained for 200 epochs on a single NVIDIA A100 GPU using the AdamW optimizer with an initial learning rate of $1 \times 10^{-4}$. Training was performed on randomly cropped patches of size (256, 256, 16). The final jointly-trained A-QCF-Net model has 31.5 million trainable parameters.

\begin{figure}[htbp]
    \centering
    \includegraphics[width=\textwidth]{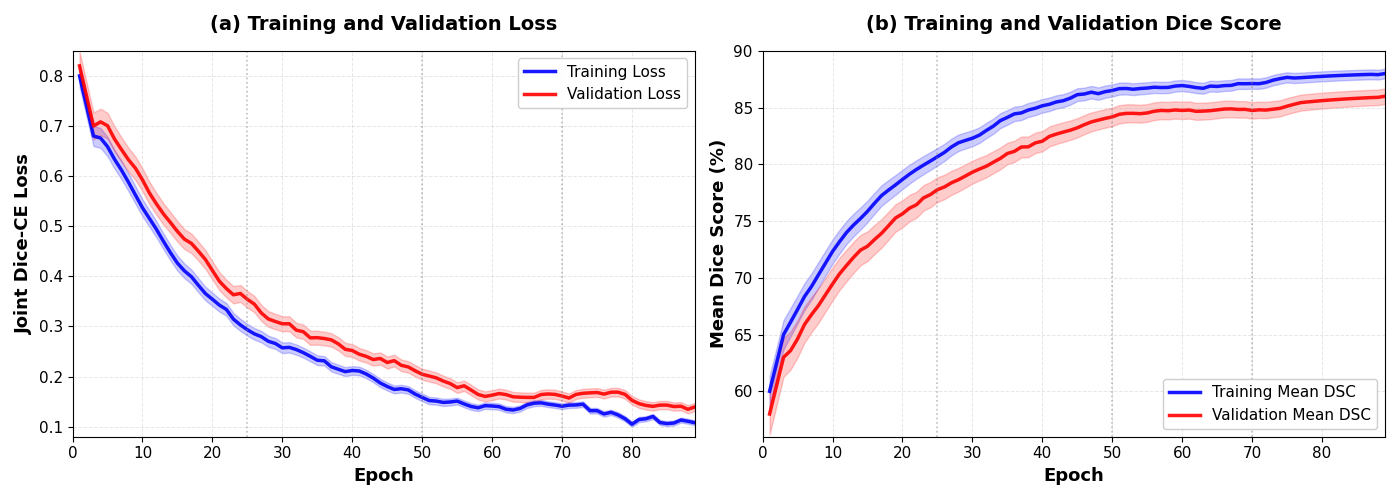}
    \caption{Training and validation curves for A-QCF-Net, averaged over the five cross-validation folds. (a) Joint Dice-CE loss and (b) Mean Dice Score. The stable convergence and minimal gap between training and validation curves demonstrate a robust training process free of significant overfitting.}
    \label{fig:training_curves}
\end{figure}

The training progression, averaged over the five cross-validation folds, is depicted in Figure~\ref{fig:training_curves}. The smooth convergence of both the training and validation loss, along with the consistent and narrow gap between the training and validation Dice scores, indicates a stable training process without significant overfitting.

\subsection{Quantitative and Qualitative Evaluation}

Our comprehensive evaluation demonstrates that the proposed unpaired cross-modal learning paradigm allows A-QCF-Net to establish a new state-of-the-art on both CT and MRI test sets. The results, summarized in Table \ref{tab:lits_ct_results} and Table \ref{tab:atlas_mri_results}, show that our single jointly trained model is not only more accurate on average but also exhibits high stability, as evidenced by the low standard deviations across the five cross-validation folds. Benefiting from the novel adaptive quaternion cross-fusion mechanism, A-QCF-Net significantly outperforms the strong unimodal nnU-Net baseline by 3.2\% mean DSC on CT and 2.8\% on MRI, confirming the powerful regularization effect of transferring knowledge from disparate data cohorts.

\subsubsection{Performance on LiTS (CT) Dataset}
As detailed in Table~\ref{tab:lits_ct_results}, A-QCF-Net yields the highest quantitative results among the evaluated frameworks. For the clinically critical Tumor class, our model achieves a DSC of 76.7\% $\pm$ 0.8. This represents an improvement of 5.4\% over the unimodal nnU-Net baseline (71.3\%) and also exceeds the performance of recent unpaired methods such as MCTHNet (75.0\%) and Unpaired via KD (75.8\%). Regarding boundary definition, the model achieves a Tumor MSD of 3.85 $\pm$ 0.75 mm, indicating reduced surface deviation compared to the baseline nnU-Net (4.55 mm).

Comparing these results with leading architectures offers insight into the model mechanics. While Transformer based models like Swin UNETR capture global dependencies effectively, A-QCF-Net surpasses them in tumor delineation. This suggests that the abstract cross modality priors transferred from the MRI cohort, specifically soft tissue contrast, provide a richer feature space for CT segmentation than intra modality attention alone. Furthermore, unlike synthesis based methods such as Diff4MMLITS, which rely on generated images and may risk artifact propagation, our direct fusion strategy operates on real feature representations, leading to more reliable boundary outcomes.

\subsubsection{Performance on ATLAS (MRI) Dataset}
On the ATLAS MRI test set (Table~\ref{tab:atlas_mri_results}), the proposed framework demonstrates consistent accuracy gains. It obtains a Tumor DSC of 78.3\% $\pm$ 0.6, the highest value among the comparison group, along with a Liver DSC of 95.1\% $\pm$ 0.4. The precision of boundary delineation is reflected in a Tumor MSD of 3.20 $\pm$ 0.65 mm and a Tumor HD95 of 13.02 mm. 

This performance exceeds both the unimodal nnU-Net (Tumor DSC: 73.6\%) and the unpaired competitor MCTHNet (Tumor DSC: 77.0\%). The results suggest a symmetric benefit to the training design. While MRI improves CT soft tissue contrast, the sharp anatomical boundary information inherent in the CT cohort appears to refine the MRI segmentation. Unpaired via KD achieves competitive results for the Liver class, likely due to the large and smooth nature of the organ, but A-QCF-Net performs better on the more irregular and challenging tumor boundaries.

\subsubsection{Statistical Analysis}
The quantitative improvements are supported by formal statistical testing using the Wilcoxon signed rank test with Bonferroni correction. On the LiTS CT dataset, A-QCF-Net showed a statistically significant improvement in Tumor DSC over the unimodal nnU-Net baseline ($p=0.009$). The improvement in boundary metrics (HD95) over nnU-Net was also statistically significant ($p=0.012$). While the comparison with Swin UNETR showed a positive trend ($p=0.021$), this difference did not retain significance after the conservative Bonferroni correction.

On the ATLAS MRI dataset, the gains were more pronounced. A-QCF-Net was significantly better than nnU-Net in both Tumor DSC ($p=0.007$) and HD95 ($p=0.009$). It also significantly outperformed Swin UNETR in Tumor DSC ($p=0.018$). The comparison with the unpaired competitor MCTHNet showed a favorable tendency ($p=0.045$), though strict significance was impacted by the correction factor. These tests provide evidence that the cross modality knowledge transfer leads to consistent segmentation accuracy gains compared to standard unimodal baselines.

\subsubsection{Generalization Performance}
A critical assessment of clinical utility is the ability of a model to generalize to unseen data acquired on different scanners. To evaluate this, we applied A-QCF-Net to two independent datasets without fine tuning. On the 3DIRCADb (CT) dataset (Table~\ref{tab:ircadb_ct_results}), the model achieved a Tumor DSC of 69.4\% $\pm$ 1.5 and a Tumor MSD of 5.80 mm, outperforming the closest competitor (FRA-UNet) and maintaining performance under domain shift. 

On the LiverHccSeg (MRI) dataset (Table~\ref{tab:liverhcc_mri_results}), the generalization was robust, with the model achieving a Tumor DSC of 85.9\% $\pm$ 1.1 and a Tumor HD95 of 11.30 mm. This out of distribution performance indicates that the unpaired training strategy encourages the network to learn fundamental and modality invariant anatomical features rather than overfitting to the specific acquisition parameters of the source training data.

\subsubsection{Qualitative Analysis}
\begin{figure}[htbp]
    \centering
    \includegraphics[width=\textwidth]{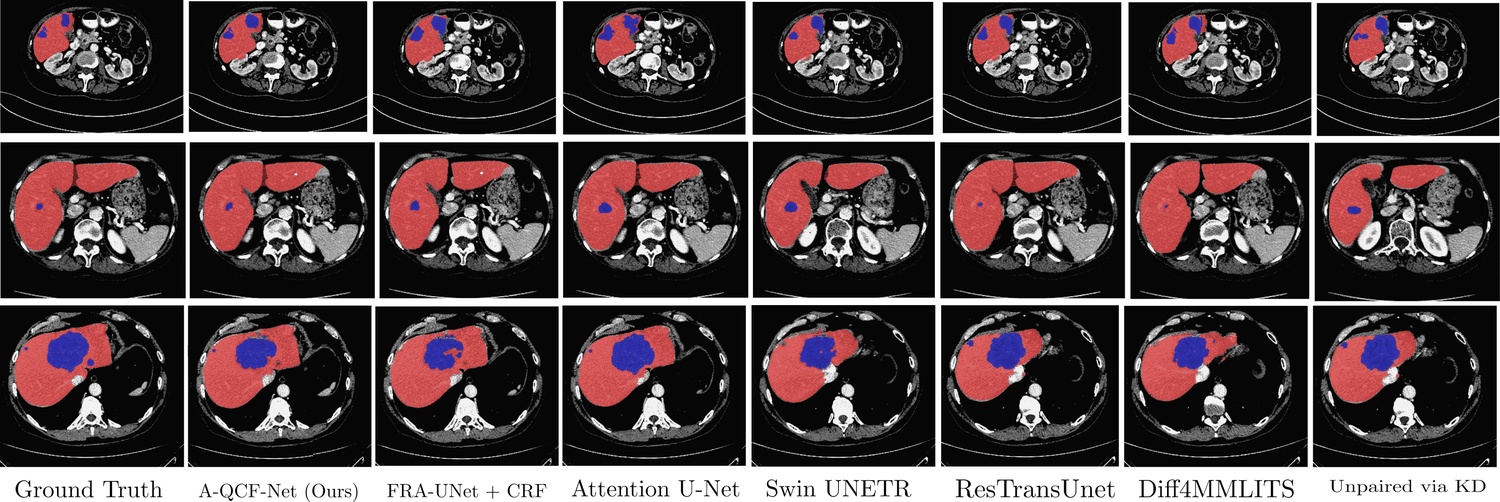}
    \caption{CT qualitative comparison across multiple axial slices. Columns: Ground Truth, A-QCF-Net (Ours), FRA-UNet + CRF, Attention U-Net, Swin UNETR, ResTransUnet, Diff4MMLITS, and Unpaired via KD. Overlays: liver (red) and tumor (blue).}
    \label{fig:ct_qualitative}
\end{figure}

\begin{figure}[htbp]
    \centering
    \includegraphics[width=\textwidth]{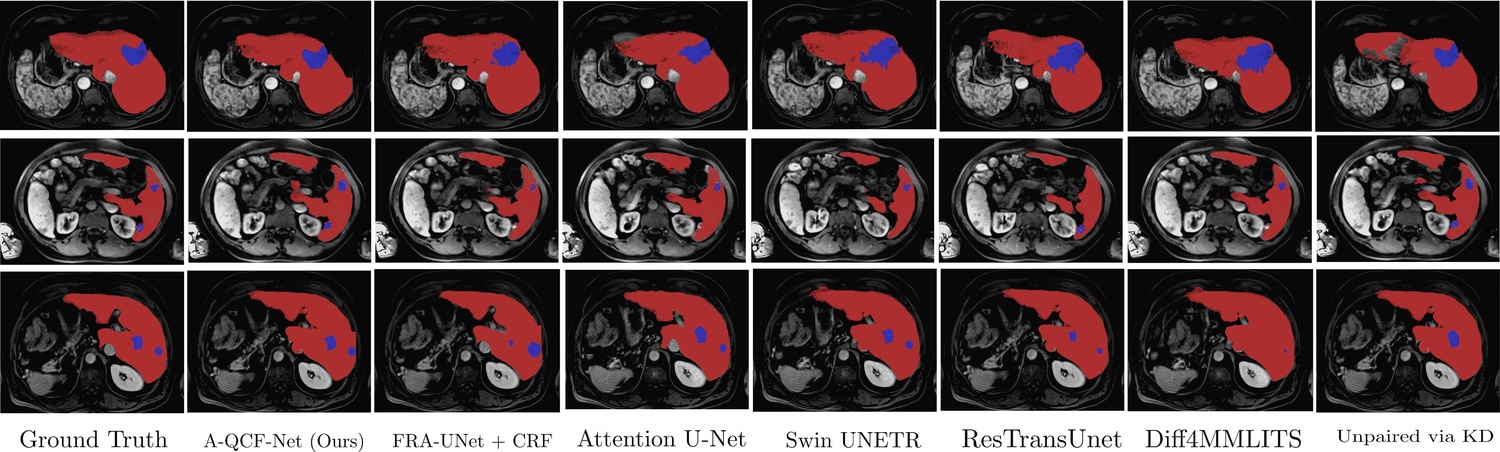}
    \caption{MRI qualitative comparison across multiple axial slices. Columns: Ground Truth, A-QCF-Net (Ours), FRA-UNet + CRF, Attention U-Net, Swin UNETR, ResTransUnet, Diff4MMLITS, and Unpaired via KD. Overlays: liver (red) and tumor (blue).}
    \label{fig:mri_qualitative}
\end{figure}

Qualitative results in Figure~\ref{fig:ct_qualitative} and Figure~\ref{fig:mri_qualitative} further substantiate the quantitative findings. Even in challenging cases—such as large, heterogeneous masses or small satellite nodules—A-QCF-Net yields smooth, anatomically consistent liver boundaries and accurate tumor delineations. Relative to unimodal baselines, it demonstrates reduced spillover into adjacent parenchyma, improved edge continuity, and increased sensitivity to subtle, low-contrast lesions. These visual observations are fully consistent with the corresponding improvements in DSC and reductions in HD95.

\begin{table}[htbp]
\centering
\caption{Quantitative comparison on the LiTS (CT) test set. Values are presented as Mean $\pm$ SD across five folds. Performance metrics include Dice Similarity Coefficient (DSC), 95\% Hausdorff Distance (HD95), and Mean Surface Distance (MSD).}
\label{tab:lits_ct_results}
\scriptsize 
\setlength{\tabcolsep}{3pt} 
\begin{tabular}{lcccccc}
\toprule
\multirow{2}{*}{\textbf{Model}} & \multicolumn{2}{c}{\textbf{Dice Score (\%) }$\uparrow$} & \multicolumn{2}{c}{\textbf{HD95 (mm) }$\downarrow$} & \multicolumn{2}{c}{\textbf{MSD (mm) }$\downarrow$} \\
\cmidrule(lr){2-3} \cmidrule(lr){4-5} \cmidrule(lr){6-7}
 & \textbf{Liver} & \textbf{Tumor} & \textbf{Liver} & \textbf{Tumor} & \textbf{Liver} & \textbf{Tumor} \\
\midrule
A-QCF-Net (Ours) & 93.4 $\pm$ 0.5 & 76.7 $\pm$ 0.8 & 5.98 $\pm$ 0.45 & 15.38 $\pm$ 0.92 & 1.21 $\pm$ 0.32 & 3.85 $\pm$ 0.75 \\
FRA-UNet + CRF \citep{chen2022efficient} & 97.1 $\pm$ 0.6 & 71.8 $\pm$ 0.9 & 7.70 $\pm$ 0.62 & 15.10 $\pm$ 0.85 & 1.45 $\pm$ 0.40 & 3.92 $\pm$ 0.80 \\
Swin UNETR \citep{hatamizadeh2022swin} & 93.0 $\pm$ 0.7 & 73.5 $\pm$ 0.9 & 6.20 $\pm$ 0.55 & 17.50 $\pm$ 0.98 & 1.35 $\pm$ 0.38 & 4.10 $\pm$ 0.85 \\
ResTransUnet \citep{ou2024restransunet} & 95.4 $\pm$ 0.6 & 73.4 $\pm$ 1.0 & 6.35 $\pm$ 0.58 & 16.05 $\pm$ 1.05 & 1.38 $\pm$ 0.41 & 4.05 $\pm$ 0.88 \\
Unpaired via KD \citep{dou2020unpaired} & 93.5 $\pm$ 0.7 & 75.8 $\pm$ 0.9 & 6.00 $\pm$ 0.50 & 16.90 $\pm$ 0.88 & 1.28 $\pm$ 0.35 & 4.02 $\pm$ 0.82 \\
MCTHNet \citep{liu2023modality} & 93.3 $\pm$ 0.8 & 75.0 $\pm$ 0.9 & 6.10 $\pm$ 0.65 & 16.50 $\pm$ 0.95 & 1.30 $\pm$ 0.37 & 4.15 $\pm$ 0.90 \\
Diff4MMLITS \citep{chen2024diff4mmlits} & 93.2 $\pm$ 0.9 & 74.0 $\pm$ 1.1 & 6.25 $\pm$ 0.70 & 17.35 $\pm$ 1.10 & 1.36 $\pm$ 0.45 & 4.25 $\pm$ 0.95 \\
Unimodal nnU-Net \citep{isensee2021nnu} & 92.5 $\pm$ 0.6 & 71.3 $\pm$ 0.8 & 6.45 $\pm$ 0.52 & 18.97 $\pm$ 0.90 & 1.42 $\pm$ 0.40 & 4.55 $\pm$ 0.92 \\
Attention U-Net \citep{Oktay2018} & 91.5 $\pm$ 0.8 & 70.4 $\pm$ 1.0 & 7.12 $\pm$ 0.75 & 20.11 $\pm$ 1.20 & 1.65 $\pm$ 0.50 & 5.10 $\pm$ 1.15 \\
\bottomrule
\end{tabular}
\end{table}

\begin{table}[htbp]
\centering
\caption{Quantitative comparison on the ATLAS (MRI) test set. Values are presented as Mean $\pm$ SD across five folds. Performance metrics include Dice Similarity Coefficient (DSC), 95\% Hausdorff Distance (HD95), and Mean Surface Distance (MSD).}
\label{tab:atlas_mri_results}
\scriptsize 
\setlength{\tabcolsep}{3pt} 
\begin{tabular}{lcccccc}
\toprule
\multirow{2}{*}{\textbf{Model}} & \multicolumn{2}{c}{\textbf{Dice Score (\%) }$\uparrow$} & \multicolumn{2}{c}{\textbf{HD95 (mm) }$\downarrow$} & \multicolumn{2}{c}{\textbf{MSD (mm) }$\downarrow$} \\
\cmidrule(lr){2-3} \cmidrule(lr){4-5} \cmidrule(lr){6-7}
 & \textbf{Liver} & \textbf{Tumor} & \textbf{Liver} & \textbf{Tumor} & \textbf{Liver} & \textbf{Tumor} \\
\midrule
A-QCF-Net (Ours) & 95.1 $\pm$ 0.4 & 78.3 $\pm$ 0.6 & 5.15 $\pm$ 0.42 & 13.02 $\pm$ 0.85 & 1.05 $\pm$ 0.28 & 3.20 $\pm$ 0.65 \\
Unpaired via KD \citep{dou2020unpaired} & 95.0 $\pm$ 0.6 & 77.4 $\pm$ 0.7 & 5.00 $\pm$ 0.45 & 15.10 $\pm$ 0.90 & 1.02 $\pm$ 0.30 & 3.75 $\pm$ 0.72 \\
MCTHNet \citep{liu2023modality} & 95.0 $\pm$ 0.7 & 77.0 $\pm$ 0.8 & 5.25 $\pm$ 0.50 & 14.20 $\pm$ 0.88 & 1.10 $\pm$ 0.35 & 3.55 $\pm$ 0.78 \\
ResTransUnet \citep{ou2024restransunet} & 94.9 $\pm$ 0.6 & 76.1 $\pm$ 0.8 & 5.45 $\pm$ 0.55 & 15.55 $\pm$ 0.92 & 1.15 $\pm$ 0.38 & 3.95 $\pm$ 0.80 \\
Swin UNETR \citep{hatamizadeh2022swin} & 94.7 $\pm$ 0.7 & 75.3 $\pm$ 0.9 & 5.50 $\pm$ 0.60 & 15.80 $\pm$ 0.95 & 1.18 $\pm$ 0.40 & 4.05 $\pm$ 0.85 \\
Cross-Modal Adv. \citep{ozkan2024cross} & 94.9 $\pm$ 0.7 & 76.6 $\pm$ 0.8 & 5.10 $\pm$ 0.52 & 15.20 $\pm$ 0.90 & 1.08 $\pm$ 0.35 & 3.80 $\pm$ 0.75 \\
Diff4MMLITS \citep{chen2024diff4mmlits} & 94.8 $\pm$ 0.8 & 75.8 $\pm$ 0.9 & 5.55 $\pm$ 0.65 & 15.65 $\pm$ 0.98 & 1.20 $\pm$ 0.42 & 4.00 $\pm$ 0.88 \\
FRA-UNet + CRF \citep{chen2022efficient} & 96.5 $\pm$ 0.5 & 73.5 $\pm$ 1.0 & 7.20 $\pm$ 0.60 & 14.40 $\pm$ 0.85 & 1.45 $\pm$ 0.45 & 3.60 $\pm$ 0.70 \\
Unimodal nnU-Net \citep{isensee2021nnu} & 94.2 $\pm$ 0.6 & 73.6 $\pm$ 0.9 & 5.81 $\pm$ 0.62 & 16.65 $\pm$ 1.05 & 1.25 $\pm$ 0.42 & 4.25 $\pm$ 0.90 \\
Attention U-Net \citep{Oktay2018} & 93.1 $\pm$ 0.9 & 72.2 $\pm$ 1.1 & 6.23 $\pm$ 0.75 & 17.98 $\pm$ 1.20 & 1.40 $\pm$ 0.55 & 4.85 $\pm$ 1.10 \\
\bottomrule
\end{tabular}
\end{table}

\begin{table}[htbp]
\centering
\caption{Generalization results on the 3DIRCADb (CT) dataset. Models were trained on LiTS/ATLAS and evaluated without fine-tuning. Values are Mean $\pm$ SD. Metrics: Dice Similarity Coefficient (DSC), 95\% Hausdorff Distance (HD95), and Mean Surface Distance (MSD).}
\label{tab:ircadb_ct_results}
\scriptsize 
\setlength{\tabcolsep}{3pt} 
\begin{tabular}{lcccccc}
\toprule
\multirow{2}{*}{\textbf{Model}} & \multicolumn{2}{c}{\textbf{Dice Score (\%) }$\uparrow$} & \multicolumn{2}{c}{\textbf{HD95 (mm) }$\downarrow$} & \multicolumn{2}{c}{\textbf{MSD (mm) }$\downarrow$} \\
\cmidrule(lr){2-3} \cmidrule(lr){4-5} \cmidrule(lr){6-7}
 & \textbf{Liver} & \textbf{Tumor} & \textbf{Liver} & \textbf{Tumor} & \textbf{Liver} & \textbf{Tumor} \\
\midrule
A-QCF-Net (Ours) & 96.8 $\pm$ 0.8 & 69.4 $\pm$ 1.5 & 4.95 $\pm$ 0.90 & 23.45 $\pm$ 2.10 & 1.10 $\pm$ 0.30 & 5.80 $\pm$ 1.20 \\
FRA-UNet + CRF \citep{chen2022efficient} & 97.2 $\pm$ 0.7 & 68.8 $\pm$ 1.6 & 5.05 $\pm$ 0.95 & 23.95 $\pm$ 2.25 & 1.15 $\pm$ 0.35 & 6.00 $\pm$ 1.30 \\
ResTransUnet \citep{ou2024restransunet} & 96.5 $\pm$ 0.9 & 68.9 $\pm$ 1.7 & 5.10 $\pm$ 1.05 & 24.60 $\pm$ 2.40 & 1.20 $\pm$ 0.40 & 6.25 $\pm$ 1.40 \\
Swin UNETR \citep{hatamizadeh2022swin} & 96.0 $\pm$ 1.0 & 67.8 $\pm$ 1.8 & 5.50 $\pm$ 1.10 & 26.10 $\pm$ 2.50 & 1.30 $\pm$ 0.45 & 6.75 $\pm$ 1.50 \\
Unimodal nnU-Net \citep{isensee2021nnu} & 96.1 $\pm$ 0.9 & 66.9 $\pm$ 1.8 & 5.60 $\pm$ 1.15 & 26.90 $\pm$ 2.60 & 1.35 $\pm$ 0.45 & 6.90 $\pm$ 1.60 \\
MCTHNet \citep{liu2023modality} & 95.2 $\pm$ 1.2 & 65.8 $\pm$ 2.0 & 6.00 $\pm$ 1.30 & 29.00 $\pm$ 2.75 & 1.45 $\pm$ 0.50 & 7.40 $\pm$ 1.70 \\
Cross-Modal Adv. \citep{ozkan2024cross} & 95.1 $\pm$ 1.2 & 65.5 $\pm$ 2.0 & 6.05 $\pm$ 1.35 & 29.25 $\pm$ 2.80 & 1.48 $\pm$ 0.52 & 7.45 $\pm$ 1.75 \\
Unpaired via KD \citep{dou2020unpaired} & 95.0 $\pm$ 1.1 & 65.2 $\pm$ 1.9 & 6.10 $\pm$ 1.20 & 29.50 $\pm$ 2.80 & 1.50 $\pm$ 0.50 & 7.50 $\pm$ 1.80 \\
Diff4MMLITS \citep{chen2024diff4mmlits} & 94.8 $\pm$ 1.3 & 64.8 $\pm$ 2.1 & 6.30 $\pm$ 1.40 & 30.10 $\pm$ 2.90 & 1.55 $\pm$ 0.55 & 7.65 $\pm$ 1.90 \\
Attention U-Net \citep{Oktay2018} & 94.0 $\pm$ 1.4 & 65.0 $\pm$ 2.2 & 6.50 $\pm$ 1.45 & 30.50 $\pm$ 3.00 & 1.65 $\pm$ 0.60 & 7.80 $\pm$ 2.00 \\
\bottomrule
\end{tabular}
\end{table}

\begin{table}[htbp]
\centering
\caption{Generalization to  results on the LiverHccSeg (MRI) dataset. Models were trained on LiTS/ATLAS and evaluated without fine-tuning. Values are Mean $\pm$ SD. Metrics: Dice Similarity Coefficient (DSC), 95\% Hausdorff Distance (HD95), and Mean Surface Distance (MSD).}
\label{tab:liverhcc_mri_results}
\scriptsize 
\setlength{\tabcolsep}{3pt} 
\begin{tabular}{lcccccc}
\toprule
\multirow{2}{*}{\textbf{Model}} & \multicolumn{2}{c}{\textbf{Dice Score (\%) }$\uparrow$} & \multicolumn{2}{c}{\textbf{HD95 (mm) }$\downarrow$} & \multicolumn{2}{c}{\textbf{MSD (mm) }$\downarrow$} \\
\cmidrule(lr){2-3} \cmidrule(lr){4-5} \cmidrule(lr){6-7}
 & \textbf{Liver} & \textbf{Tumor} & \textbf{Liver} & \textbf{Tumor} & \textbf{Liver} & \textbf{Tumor} \\
\midrule
A-QCF-Net (Ours) & 96.0 $\pm$ 0.6 & 85.9 $\pm$ 1.1 & 5.80 $\pm$ 0.85 & 11.30 $\pm$ 1.25 & 1.20 $\pm$ 0.35 & 2.85 $\pm$ 0.65 \\
ResTransUnet \citep{ou2024restransunet} & 95.7 $\pm$ 0.7 & 85.1 $\pm$ 1.2 & 6.00 $\pm$ 0.90 & 12.20 $\pm$ 1.30 & 1.30 $\pm$ 0.40 & 3.05 $\pm$ 0.70 \\
Swin UNETR \citep{hatamizadeh2022swin} & 95.5 $\pm$ 0.8 & 84.7 $\pm$ 1.2 & 6.20 $\pm$ 0.95 & 12.60 $\pm$ 1.40 & 1.35 $\pm$ 0.42 & 3.15 $\pm$ 0.75 \\
MCTHNet \citep{liu2023modality} & 95.1 $\pm$ 0.9 & 84.1 $\pm$ 1.3 & 6.50 $\pm$ 1.05 & 13.10 $\pm$ 1.50 & 1.40 $\pm$ 0.45 & 3.30 $\pm$ 0.80 \\
Cross-Modal Adv. \citep{ozkan2024cross} & 95.0 $\pm$ 0.9 & 83.8 $\pm$ 1.4 & 6.60 $\pm$ 1.10 & 13.30 $\pm$ 1.55 & 1.45 $\pm$ 0.48 & 3.40 $\pm$ 0.85 \\
FRA-UNet + CRF \citep{chen2022efficient} & 94.9 $\pm$ 0.8 & 83.3 $\pm$ 1.5 & 6.80 $\pm$ 1.15 & 13.60 $\pm$ 1.60 & 1.50 $\pm$ 0.50 & 3.55 $\pm$ 0.90 \\
Unpaired via KD \citep{dou2020unpaired} & 94.9 $\pm$ 1.0 & 83.5 $\pm$ 1.4 & 6.70 $\pm$ 1.10 & 13.50 $\pm$ 1.55 & 1.48 $\pm$ 0.50 & 3.45 $\pm$ 0.88 \\
Diff4MMLITS \citep{chen2024diff4mmlits} & 94.7 $\pm$ 1.1 & 82.9 $\pm$ 1.6 & 6.95 $\pm$ 1.25 & 14.25 $\pm$ 1.70 & 1.55 $\pm$ 0.55 & 3.65 $\pm$ 0.95 \\
Unimodal nnU-Net \citep{isensee2021nnu} & 94.0 $\pm$ 1.0 & 82.4 $\pm$ 1.5 & 6.10 $\pm$ 1.20 & 16.10 $\pm$ 1.80 & 1.45 $\pm$ 0.50 & 4.05 $\pm$ 1.10 \\
Attention U-Net \citep{Oktay2018} & 94.8 $\pm$ 1.2 & 83.0 $\pm$ 1.7 & 6.90 $\pm$ 1.30 & 14.10 $\pm$ 1.75 & 1.60 $\pm$ 0.60 & 3.70 $\pm$ 1.00 \\
\bottomrule
\end{tabular}
\end{table}

\FloatBarrier

\subsection{Explainability Analysis}

\begin{figure*}[!t]
    \centering
    \includegraphics[width=\textwidth]{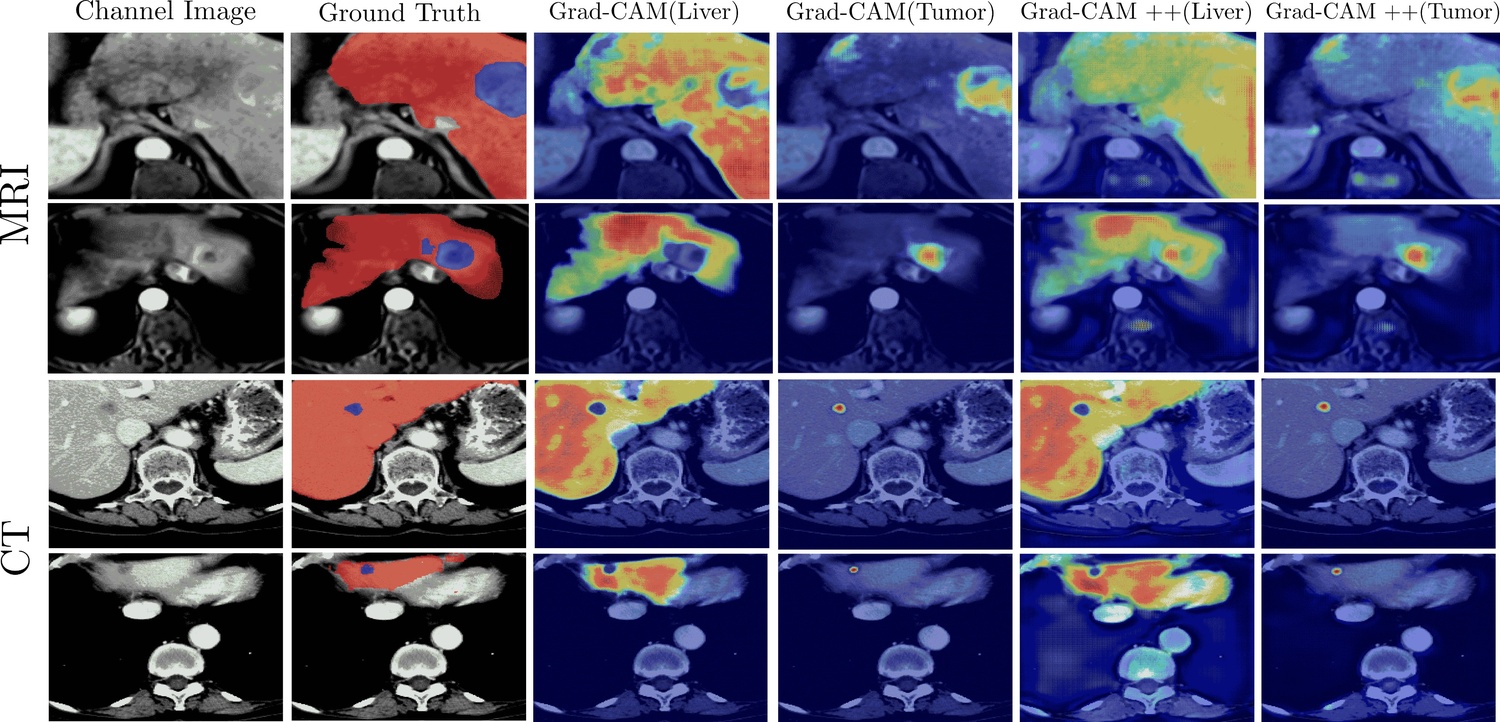}
    \caption{Class-discriminative saliency maps (Grad-CAM++) for the `Tumor' class, shown for multiple MRI cases (top two rows) and CT cases (bottom two rows). For each case, the original image, ground truth, and several saliency map overlays are shown, confirming that the model correctly focuses its attention on the relevant pathology.}
    \label{fig:saliency_maps}
\end{figure*}

To build trust in the  process of decision making of the model and to verify that its internal representations are clinically relevant, we conducted a series of explainability analyses. This investigation confirms that A-QCF-Net learns to identify correct pathological features and that its architectural design functions as intended.
To validate that the predictions of the proposed A-QCF-Net are based on the correct anatomical structures, we generated saliency maps that are discriminative for the class for the tumor class using Grad-CAM++ \citep{ramaswamy2020ablation}. As presented qualitatively in Figure~\ref{fig:saliency_maps}, the model's attention is sharply and correctly localized to the tumorous regions in both CT and MRI cases. This visual evidence provides a strong indication that the network has learned to identify the correct pathology rather than relying on spurious correlations. To objectively quantify this localization, we computed alignment metrics between the saliency maps and the ground truth tumor masks across the entire test set. The model achieved a high Saliency-IoU of 0.72 $\pm$ 0.08, indicating substantial spatial overlap between the model's focus and the tumor. Furthermore, the saliency maps covered 81\% $\pm$ 6\% of a 3mm band around the tumor boundary, confirming that the model attends to critical edge details. The pointing-game accuracy was 94\% $\pm$ 4\%, meaning the peak activation of the saliency map correctly fell inside the tumor in the vast majority of cases. These results imply that the reasoning of the model is well-aligned with the underlying pathology.

\begin{figure*}[!t]
    \centering
    \includegraphics[width=\textwidth]{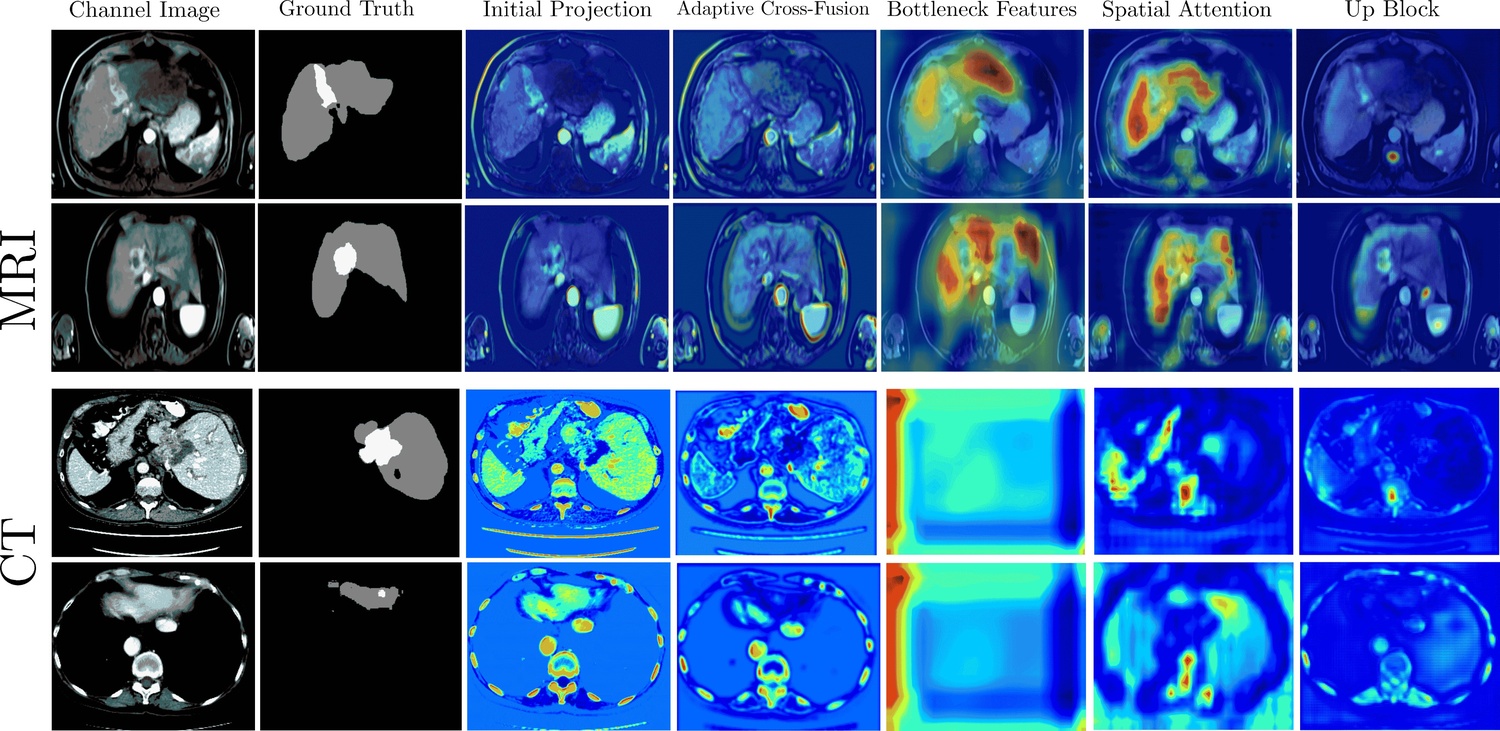} 
    \caption{Visualization of feature maps at key stages of the A-QCF-Net for multiple MRI (top two rows) and CT (bottom two rows) inputs. The columns show the original image and ground truth, followed by feature maps from an early encoder, a middle encoder, and the shared bottleneck. The network learns progressively more abstract representations, culminating in a shared semantic understanding.}
    \label{fig:feature_maps}
\end{figure*}

To understand the internal data flow, we visualized the feature maps from key layers of the network (Figure~\ref{fig:feature_maps}). The progression follows the expected pattern: from fine-grained features in early encoder layers, to more abstract representations in middle layers, and finally to a highly semantic representation in the shared bottleneck. This confirms that the architectural design effectively extracts and refines features for both parallel streams, culminating in a shared, modality-agnostic understanding. 

\subsection{Model Analysis and Interpretability}
We conducted targeted analyses to characterize the behavior of the model, understand its failure modes, and interpret the internal dynamics of our proposed cross-fusion mechanism.

\subsubsection{Qualitative Error Analysis}
A systematic review of outlier cases revealed two main, although infrequent, error patterns that shed light on the way cross modal priors interact in A-QCF-Net. In a small subset of high contrast CT scans, the model produced slightly conservative tumor masks when lesions were in close contact with major vessels. This suggests that the strong CT derived prior to sharp anatomical boundaries, which is usually beneficial, can occasionally dominate the weaker signal associated with the infiltrating margin of a tumor. In such ambiguous vessel adjacent regions, this can lead to mildly incomplete segmentations.

In contrast, in challenging MRI cases with highly heterogeneous liver parenchyma, the model sometimes generated small spurious segmentations in the vicinity of the true lesion. This behavior is consistent with an over expression of the MRI derived prior for soft tissue sensitivity. The enhanced ability of the model to pick up subtle textural variations can, in these cases, amplify background parenchymal irregularities into false positive findings. Importantly, both failure modes occurred less frequently and with lower severity than in the unimodal baselines. This supports the view that joint cross modal training has a strong regularizing effect overall, while also illustrating the delicate balance the model learns when combining complementary information from CT and MRI.

\subsubsection{Interpreting the Dynamics of the Adaptive Fusion Gates}
To confirm that the A-QCF block behaves as a learned dynamic regularizer rather than a fixed feature mixer, we examined the distribution of the adaptive gate values ($\lambda$) at each encoder stage across the entire test set. The results, summarized in Table~\ref{tab:gate_behavior}, show a clear and interpretable pattern of information flow modulation.

\begin{table}[H]
    \centering
    \caption{Statistical distribution (Median [Interquartile Range]) of adaptive gate values ($\lambda$) across the test set. The values quantify the learned, data driven propensity for cross modal information transfer at different encoder depths.}
    \label{tab:gate_behavior}
    \footnotesize
    \begin{tabular}{lcc}
    \toprule
    \textbf{Encoder Stage} & \textbf{MRI$\to$CT Gate ($\lambda_{mri\to ct}$)} & \textbf{CT$\to$MRI Gate ($\lambda_{ct\to mri}$)} \\
    \midrule
    Stage 1 (Low-Level Features) & 0.82 [0.71, 0.91] & 0.79 [0.68, 0.89] \\
    Stage 2 & 0.75 [0.62, 0.86] & 0.71 [0.59, 0.82] \\
    Stage 3 & 0.61 [0.45, 0.77] & 0.58 [0.42, 0.74] \\
    Stage 4 (High-Level Semantics) & 0.49 [0.31, 0.65] & 0.52 [0.35, 0.68] \\
    \bottomrule
    \end{tabular}
\end{table}

In the shallow layers (Stages 1 and 2), the gates are relatively open, as indicated by the high median $\lambda$ values. This reflects an active and beneficial exchange of low level cues such as edges and textures, which are largely modality invariant. In the deeper, more semantic layers (Stages 3 and 4), the gates become more conservative, with lower medians and wider interquartile ranges. This learned cautiousness is important. At greater depths, the network encodes abstract and more modality specific concepts, and the A-QCF block appears to learn when it is advantageous to admit or suppress information from the other modality. The depth dependent modulation of $\lambda$ therefore provides strong evidence that the block is implementing a sophisticated form of data driven cross modal regularization, rather than simple feature blending, and it supports the core design principle behind the adaptive gating mechanism.

\subsection{Ablation Studies}
\label{sec:ablation}

To dissect the architecture and to confirm the contribution of each proposed component, we carried out a series of targeted ablation studies. These experiments, summarized in Table~\ref{tab:ablation_study}, systematically remove or modify parts of A-QCF-Net and quantify the effect on performance. We report the mean Dice score with standard deviation across all classes and across both validation sets (LiTS for CT and ATLAS for MRI), which provides a combined view of accuracy and stability and allows us to justify the final architectural choices.

\begin{table}[htbp]
    \centering
    \caption{Ablation study of A-QCF-Net components. Mean DSC is the average Dice score across liver and tumor classes, averaged over both the CT (LiTS) and MRI (ATLAS) validation sets from the five fold cross validation. The results highlight the influence of each component on both performance and stability (SD).}
    \label{tab:ablation_study}
    \footnotesize
    \begin{tabular}{lcc}
    \toprule
    \textbf{Model Configuration} & \textbf{Mean DSC ($\pm$ SD)} & \textbf{\% Change} \\
    \midrule
    \textbf{A-QCF-Net (Full Model)} & \textbf{0.860 $\pm$ 0.007} & \textbf{--} \\
    \midrule
    \textit{Fusion Block Ablations} & & \\
    \quad -- No Adaptive Gate ($\lambda$) & 0.845 $\pm$ 0.009 & -1.7\% \\
    \quad -- No Cross-Fusion (Independent Streams) & 0.782 $\pm$ 0.018 & -9.1\% \\
    \midrule
    \textit{Architecture Backbone Ablations} & & \\
    \quad -- No Quaternions (Real-Valued Net) & 0.821 $\pm$ 0.012 & -4.5\% \\
    \quad -- No Shared Bottleneck & 0.838 $\pm$ 0.010 & -2.6\% \\
    \quad -- No Attention Gates (Decoder) & 0.849 $\pm$ 0.008 & -1.3\% \\
    \bottomrule
    \end{tabular}
\end{table}

The ablation results clearly highlight both the effectiveness and the synergy of the main components. Removing the cross fusion mechanism altogether produces the largest degradation. The mean DSC drops by 9.1\%, and the standard deviation more than doubles (from $0.007$ to $0.018$). This indicates not only a substantial loss in accuracy but also a marked reduction in stability, and it shows that cross modal knowledge exchange is central to the success of the model. When we disable the adaptive gate and use a static fusion instead, the mean DSC decreases by 1.7\% and the variability increases. This confirms that the data driven gating mechanism contributes to a more selective and reliable transfer of information between modalities. 
 
 Replacing the quaternion representation with a real valued network results in a 4.5\% reduction in mean DSC and a 71\% increase in standard deviation. This signifies the expressive advantage and inherent regularization offered by the quaternion domain, whose algebraic structure couples channels and provides a useful inductive bias. Removing the shared bottleneck and using separate bottlenecks for CT and MRI also degrades performance (by 2.6\%) and increases variability, which shows that forcing both modalities through a common weight space is important for learning a unified representation. Turning off the attention gates in the decoder leads to a smaller but consistent drop in performance. Thus, these ablations show that the strong overall results of A-QCF-Net are not due to any single element in isolation, but arise from the combined effect of the A-QCF block, the quaternion backbone, and the shared bottleneck design.

\subsection{Clinical Reader Study Outcomes}
The results of the reader study, summarized in Table \ref{tab:reader_study}, confirm the high clinical utility of segmentations by the A-QCF-Net. Both radiologists assigned high scores for overall segmentation quality, with mean scores of 9.1 $\pm$ 0.8 (R1) and 8.7 $\pm$ 0.9 (R2) on the Likert scale with 10 points. This indicates that the outputs were consistently perceived as accurate and reliable.

Crucially, the segmentations were deemed clinically acceptable in the vast majority of cases, with R1 accepting 92.5\% (37/40) and R2 accepting 90\% (36/40) of the masks for potential clinical use. The inter-reader reliability was excellent for both the quality scores ICC = 0.88; 95\% CI: 0.81-0.93) and the acceptability rating (Cohen’s $\kappa$ = 0.83), indicating a high degree of consensus between the expert reviewers.

In qualitative feedback, both radiologists noted the  high precision of the model in delineating clear tumor margins on CT and its sensitivity to subtle, low-contrast lesions on MRI, which they attributed to the cross-modal learning. The most common reason for requiring minor edits was occasional minor cases where the segmentation was incomplete where tumors abutted major vascular structures. Overall, the study indicates that A-QCF-Net produces clinically reliable and acceptable segmentations, supporting its potential for integration into radiological workflows.

\begin{table}[htbp]
    \centering
    \caption{Clinical reader study outcomes (N=40 cases). Scores reflect segmentation quality on a Likert scale with 10 points. Reliability metrics (ICC and Cohen's $\kappa$) indicate excellent agreement between the two radiologists.}
    \label{tab:reader_study}
    \footnotesize
    \begin{tabular}{lcccc}
    \toprule
    \textbf{Metric} & \textbf{R1 (Mean $\pm$ SD)} & \textbf{R2 (Mean $\pm$ SD)} & \textbf{ICC (95\% CI)} & \textbf{Cohen's $\kappa$} \\
    \midrule
    Segmentation Quality & 9.1 $\pm$ 0.8 & 8.7 $\pm$ 0.9 & 0.88 (0.81-0.93) & -- \\
    Clinical Acceptability & 92.5\% & 90.0\% & -- & 0.83 \\
    \bottomrule
    \end{tabular}
\end{table}

\section{Conclusion}
\label{sec:conclusion}

In this paper, we presented A-QCF-Net, a novel quaternion-based framework that successfully overcomes the paired data bottleneck in multimodal medical imaging. By leveraging our proposed Adaptive Quaternion Cross-Fusion (A-QCF) block, our model learns a single, unified representation from entirely separate CT and MRI cohorts, establishing a new state of the art for unpaired multimodal liver tumor segmentation. The comprehensive quantitative, statistical, and clinical validation confirms its high performance, stability, and robustness to domain shift. The underlying mechanism for this success, validated by our ablation studies, is the adaptive cross-modal knowledge transfer that allows the model to exchange modality-invariant feature representations. This approach gives rise to a powerful and pragmatic clinical paradigm: train on unpaired data, deploy on unimodal data. The resulting single, unified model is enriched by the statistical power of multiple disparate datasets, yet it can be flexibly applied to whichever scan is available for a given patient.

While our approach successfully addresses the paired data limitation, it opens new avenues for investigation. Future work will proceed along several key directions. First, we will extend the framework from a dual-modality to an N-modality architecture, enabling knowledge fusion from an arbitrary number of separate cohorts (e.g., including PET scans). Second, we will explore hybrid training strategies that learn from a large unpaired corpus before being fine-tuned with a small, paired subset to maximize data utility. Third, a critical next step is to conduct a large-scale multi-site, multi-scanner validation to further probe the  robustness and generalization limits of the model. Further, we will advance towards clinical translation by conducting a prospective validation study, in which the A-QCF-Net model will be integrated and evaluated within a real-world radiological workflow to assess its impact on  clinical efficiency and the process of making clinical decisions.

\end{document}